\documentclass[10pt,twocolumn,letterpaper]{article}

\usepackage[pagenumbers]{cvpaper}
\usepackage{colortbl}

\usepackage{booktabs}
\usepackage{pifont}
\usepackage{xspace}
\usepackage{amsmath,amssymb,amsfonts}
\usepackage{enumitem}     % algorithm enumerate options
\usepackage{float}        % \newfloat for the algorithm environment
\usepackage{graphicx}
\usepackage{multirow}
\usepackage{tabularx}

\makeatletter
\@ifundefined{algorithm}{%
  \newfloat{algorithm}{tbp}{loa}
  \floatname{algorithm}{Algorithm}
}{}
\makeatother
\newcommand{\ours}{\emph{\textsc{Scalar}}\xspace}
\providecommand{\gram}{\textsc{Gram}}
\providecommand{\hypergram}{\textsc{HyperGram}}

\providecommand{\pmrl}{\textsc{PMRL}}

\newcommand{\cmark}{\ding{51}}
\newcommand{\xmark}{\ding{55}}

\definecolor{cvblue}{rgb}{0.21,0.49,0.74}
\usepackage[breaklinks,colorlinks,allcolors=cvblue]{hyperref}

\title{Query-Conditioned Spherical Centroid Aggregation for Multimodal Retrieval}

\author{Ambuj Mehrish \quad Anindya Nag \quad Sebastiano Vascon\\
Department of Environmental Science, Informatics and Statistics\\
Ca' Foscari University of Venice, Italy\\
{\tt\small \{ambuj.mehrish, anindya.nag, sebastiano.vascon\}@unive.it}
}
\hypersetup{
  pdftitle={Query-Conditioned Spherical Centroid Aggregation for Multimodal Retrieval},
  pdfauthor={Ambuj Mehrish, Anindya Nag, Sebastiano Vascon}
}

\begin{document}
\maketitle
\begin{abstract}
Multimodal retrieval integrates video, audio, subtitles, and text; however, recent geometric aggregators, such as Gramian volumes, hyperbolic volumes, and spectral objectives, treat all modalities symmetrically. Under a unified evaluation protocol, their joint scores frequently lag behind the strongest single-modality pathway by 1.9 to 27.6 R@1. Controlled analyses attribute this outcome to uniform modality influence. This work introduces Spherical Centroid Aggregation with Learned Adaptive Relevance (SCALAR), a query-conditioned aggregator that assigns relevance-based weights to each available modality before computing a spherical centroid. SCALAR accommodates arbitrary modality subsets and is trained on masked, reduced-arity views using rank-8 LoRA adapters. Across five benchmarks, SCALAR achieves positive aggregation gain on four, reaching +4.0 R@1, while none of the evaluated prior aggregators is positive on more than one. A uniform-weight ablation reproduces the degradation observed with symmetric aggregation. With only 4.8 million trainable parameters, SCALAR attains the highest text-to-video R@1 on three benchmarks and performs within seed variation of the best result on a fourth. Under test-time modality dropout, SCALAR's representation-stage score surpasses the released GRAM checkpoint at every evaluated masking rate and benchmark by 3.2 to 10.9 R@1. Finally, as modalities are removed, rerankers trained exclusively on complete modality sets increasingly converge toward their video-only pathways, diminishing these representation-level gains and underscoring a limitation of standard two-stage retrieval pipelines. 
\end{abstract}
    
\section{Introduction}
\label{sec:intro}
Recent advances in text-to-video retrieval increasingly employ multimodal base models~\cite{Chen2023VASTAV,Chen2023VALORVO,Girdhar2023ImageBindOE,Zhu2023LanguageBindEV} that encode video, audio, and subtitle information into a unified embedding space for comparison with text queries. This progress has established modality aggregation as a distinct research challenge. Current methods move beyond pairwise similarity~\cite{cicchetti2025gramian,you2025mover} by introducing geometric objectives over the entire modality set, including the volume of the parallelotope~\cite{cicchetti2025gramian} defined by the embeddings, its hyperbolic analogue~\cite{na2026hypergram}, and the dominant eigenvalue of the Gram matrix~\cite{liu2026principled}. The central hypothesis is that these more complex geometric forms capture cross-modal relationships that pairwise metrics may not detect. Nevertheless, these objectives are inherently symmetric, assigning each modality an equivalent structural role~\cite{jeong2024anchors,yin2026towards} regardless of its relevance to the query. This symmetry may be restrictive, as the most informative modality can vary significantly and is often model-dependent. Consequently, a multimodal aggregation rule must meet a fundamental yet challenging criterion: it should outperform the strongest unimodal score.

Combining modalities does not consistently outperform the strongest unimodal representation~\cite{wang2020makes,wu2022characterizing,huang2021makes,du2023uni}. Previous research attributes this phenomenon to modality imbalance, competition during joint training~\cite{huang2021makes}, and noise introduced by fixed fusion~\cite{Ibrahimi2023AudioEnhancedTR}. This study investigates whether a similar trend occurs in geometric aggregation by comparing each joint score with its strongest single-modality pathway using a standardized protocol. Among prior methods, the joint score is lower than the strongest single-modality
pathway for nearly every method and benchmark we measure, with deficits reaching
27.6 R@1 points (Table~\ref{tab:gain}). We define this difference as \emph{aggregation gain}: positive values indicate improvement over the strongest pathway, while negative values suggest that aggregation diminishes an otherwise informative signal.

The observed behavior is partly attributable to symmetric aggregation, which fails to adapt each modality's contribution to the query. Substituting query-conditioned weights with uniform weights decreases aggregation gain by $7.0$, $7.5$, $4.0$, and $11.2$ R@1 points on MSR-VTT~\cite{MSRVTT}, DiDeMo~\cite{DIDEMO}, ActivityNet~\cite{ACTIVITYNET}, and VATEX~\cite{wang2019vatex}, respectively. Since the uniform spherical centroid exhibits deficits similar to geometric baselines that do not utilize volumes or eigenspectra, this effect appears to extend beyond any particular geometric construction. Symmetry is less detrimental when modality pathways provide comparable information, although volume scores may still be influenced by query-independent inter-modality terms. Collectively, these results support query-dependent modality weighting.

An effective aggregation rule must remain robust as modality availability changes. This variation is evident across the benchmarks: MSR-VTT and VATEX include video, audio, and subtitles, while DiDeMo, ActivityNet, and AudioCaps utilize only video and audio. Since geometric scores are influenced by the number of embeddings, the removal of a modality alters both the available evidence and the score scale. The proposed method addresses this challenge by using a query-conditioned spherical centroid that weights each observed modality according to its relevance to the query. The resulting cosine score is defined on a unified scale for any non-empty subset of modalities, eliminating the need for imputation or an additional fusion network. A single temperature parameter regulates weight concentration, and reduced-arity training combined with rank-8 LoRA adapts only 4.8 million parameters. 

Our contributions are threefold. (i) \ours{}, a query-conditioned spherical centroid, defined as $w_m \propto \exp(\langle t, z_m\rangle/\tau_w)$, which is applicable to any non-empty modality subset and is trained using rank-8 LoRA on 4.8 million parameters. This approach outperforms full fine-tuning under an identical training recipe by an average of 2.4 R@1. (ii) Aggregation gain, defined as the R@1 difference between the joint score and the strongest single pathway, serves as a diagnostic, under which prior geometric aggregators yield negative values in nearly every measurable case (Table.~\ref{tab:gain}). (iii) We introduce a deterministic missing-modality protocol evaluated at five removal rates. Under this protocol, rerankers trained exclusively on complete modality sets converge toward their video-only pathways. Thus, representation-level robustness does not necessarily persist through two-stage retrieval (Table.~\ref{tab:missing}).

\section{Related Work}
\label{sec:related}
\noindent\textbf{Multimodal Retrieval and Geometric Aggregation.} Omni-modal models embed video, audio, subtitles, and text within a unified space for retrieval~\cite{Chen2023VASTAV,Chen2023VALORVO,Girdhar2023ImageBindOE,Zhu2023LanguageBindEV}. VAST~\cite{Chen2023VASTAV}, which serves as the backbone for this study, treats subtitles as a primary modality alongside vision and audio. Recent approaches aggregate these streams using higher-order geometric techniques: \gram{}~\cite{cicchetti2025gramian} employs Gramian volume, \hypergram{}~\cite{na2026hypergram} integrates Euclidean and hyperbolic volumes, and \pmrl{}~\cite{liu2026principled} optimizes the dominant eigenvalue of the Gram matrix. These scoring methods are symmetric with respect to the modalities and depend on their quantity. As a result, they do not consider query-specific modality relevance and do not provide a directly comparable scale across different modality subsets.

\noindent\textbf{Query-Conditioned Fusion.} Query-dependent weighting is a well-established approach in expert-fusion methods, including MoEE~\cite{miech2018learning}, Collaborative Experts~\cite{liu2019use}, MMT~\cite{gabeur2020multi}, and X-Pool~\cite{9879391}. Prior research also demonstrates that multimodal fusion may not surpass the strongest unimodal pathway when weakly informative streams introduce noise or compete during training~\cite{wang2020makes,wu2022characterizing,du2023uni}. In contrast to learned fusion networks, the proposed method derives modality weights directly from query--modality agreement.
% We evaluate its contribution using \emph{aggregation gain}, defined as the improvement in the joint score over the strongest unimodal score within the same embedding model.

\noindent\textbf{Incomplete Modalities and Spherical Aggregation.} Prior work addresses missing modalities through reconstruction, prompting, or robust representation learning~\cite{ma2021smil,lee2023multimodal,ma2022multimodal,wu2024deep}. Retrieval adds a distinct requirement: scores obtained from different observed modality subsets must remain comparable. Our approach uses a weighted spherical mean~\cite{buss2001spherical,banerjee2005clustering}, defined for every non-empty set and returning a cosine similarity on a normalized scale. This robustness is enforced at the representation stage, since a downstream reranker can only reorder candidates retrieved by the initial encoder~\cite{Li2021AlignBF,geigle2022retrieve}.

\section{Method}
\label{sec:method}

% We introduce \ours (\textbf{S}pherical \textbf{C}entroid
% \textbf{A}ggregation), a query-conditioned aggregation framework for multimodal retrieval. \ours is designed to remain well-defined when the available set of modalities changes, while allowing each modality to contribute according to its relevance to the current query. 

\subsection{Problem Setup and Motivation}
\label{sec:problem}

\begin{figure*}
    \centering
    \includegraphics[width=0.9\textwidth]{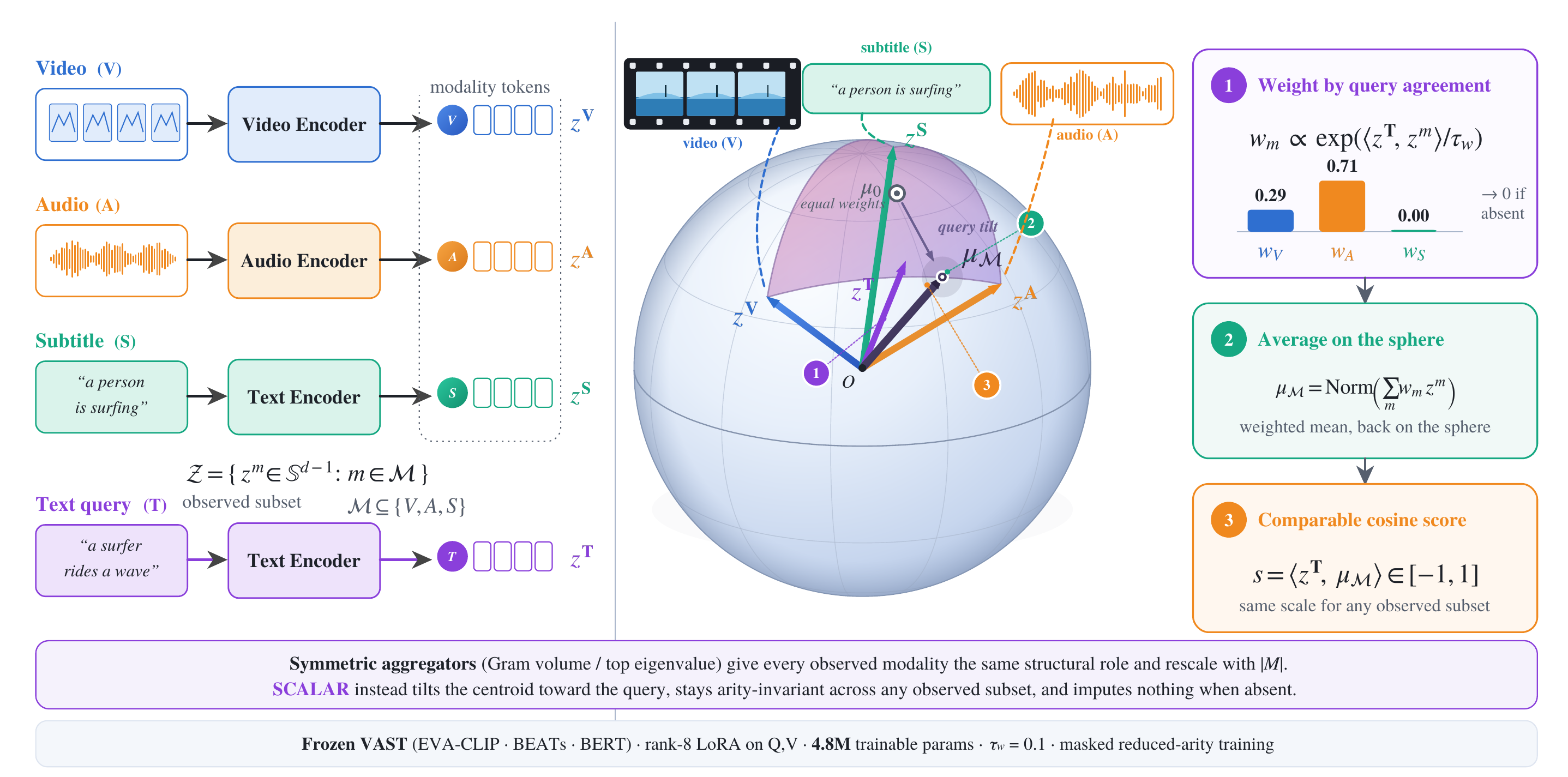}
\caption{\textbf{Overview of \ours.}
Query-dependent weights combine the observed modality embeddings into a normalized spherical centroid. Its cosine similarity to the query provides a common score for any non-empty modality subset, while absent modalities receive zero weight.}
    \label{fig:method}
\end{figure*}

We address text-based retrieval across samples that may include video, audio, and, when available, subtitles. Let \(\mathcal{M}\subseteq\{V,A,S\}\) represent the observed modalities, \(\mathcal{Z}=\{z^m\in\mathbb{S}^{d-1}:m\in\mathcal{M}\}\) their normalized embeddings, and \(z^T\in\mathbb{S}^{d-1}\) the query embedding. Our objective is to define a similarity function \(s(z^T,\mathcal{Z})\in[-1,1]\) that enables ranking of samples with varying \(|\mathcal{M}|\) without requiring arity-specific adjustments. Existing geometric scores based on the Gram matrix \(G\), such as \(\sqrt{\det(G)}\) or \(\lambda_1(G)\), capture higher-order interactions but are sensitive to the number of embeddings: the volume changes dimensionality, and for \(k\) unit vectors, \(\operatorname{tr}(G)=k\) and \(1\leq\lambda_1(G)\leq k\). Additionally, their symmetry precludes explicit query-dependent weighting, even though the most informative modality may differ across queries. \ours resolves both issues by introducing a query-conditioned spherical centroid over the available modalities, as shown in Figure~\ref{fig:method}.

\subsection{Query-Conditioned Spherical Aggregation}
\label{sec:aggregation}

The spherical centroid of the observed modalities is
\begin{equation}
    \mu_{\mathcal{M}}
    =
    \frac{\sum_{m\in\mathcal{M}} z^m}
    {\left\|\sum_{m\in\mathcal{M}} z^m\right\|}
    \in\mathbb{S}^{d-1},
    \label{eq:centroid}
\end{equation}
which is defined for any non-empty $\mathcal{M}$. Uniform averaging,
however, gives every observed modality equal influence. \ours instead
weights each modality by its agreement with the query and aggregates
accordingly,
\begin{equation}
\begin{aligned}
    w_m(z^T)
    &=
    \frac{\exp(\langle z^T,z^m\rangle/\tau_w)}
         {\sum_{m'\in\mathcal{M}}\exp(\langle z^T,z^{m'}\rangle/\tau_w)},
    \\[4pt]
    \mu_{\mathcal{M}}(z^T)
    &=
    \frac{\sum_{m\in\mathcal{M}}w_m(z^T)\,z^m}
         {\bigl\|\sum_{m\in\mathcal{M}}w_m(z^T)\,z^m\bigr\|},
\end{aligned}
\label{eq:qcentroid}
\end{equation}
where the softmax is restricted to the observed set, so an absent
modality receives exactly zero weight. The retrieval score is the
cosine
\begin{equation}
    s(z^T,\mathcal{Z})=\langle z^T,\mu_{\mathcal{M}}(z^T)\rangle .
    \label{eq:score}
\end{equation}
Because the centroid is always normalized onto the unit sphere, this
score remains in $[-1,1]$ irrespective of $|\mathcal{M}|$. The
temperature $\tau_w$ controls the degree of query specialization:
$\tau_w\!\rightarrow\!\infty$ recovers the uniform centroid of
Eq.~\eqref{eq:centroid}, whereas $\tau_w\!\rightarrow\!0$ approaches
selection of the single most query-relevant modality. We use
$\tau_w=0.1$.

\noindent\textbf{Efficient scoring.}
Although the centroid is query dependent, it need not be materialized
for every query--sample pair. Substituting
Eq.~\eqref{eq:qcentroid} into Eq.~\eqref{eq:score} gives
\begin{equation}
    s(z^T,\mathcal{Z})
    =
    \frac{\sum_{m}w_m\langle z^T,z^m\rangle}
         {\sqrt{\sum_{m,n}w_m w_n\langle z^m,z^n\rangle}},
    \label{eq:closedform}
\end{equation}
with sums over $m,n\in\mathcal{M}$. The within-sample Gram matrix
$[\langle z^m,z^n\rangle]$ is $|\mathcal{M}|\times|\mathcal{M}|$ and is
computed once per sample, so scoring never stores a $d$-dimensional centroid per query--sample pair. Algorithm~\ref{alg:supp-score} of the supplement states the resulting scoring procedure.

\subsection{Training}
\label{sec:training}

\noindent\textbf{Reduced-arity views.}
Being defined for arbitrary modality subsets does not by itself ensure
robustness to missing modalities. Let $\mathcal{K}$ denote the complete
set of modalities available for a training sample. The probability of
using the complete view is linearly annealed from $1$ to $0.5$ over the
first $2000$ steps; otherwise one modality
$m^\dagger\sim\mathrm{Unif}(\mathcal{K})$ is removed, giving
$\mathcal{M}=\mathcal{K}\setminus\{m^\dagger\}$, provided at least two
modalities remain. Masking is applied after encoding, so the complete
and reduced-arity views are obtained from the same forward pass.

\noindent\textbf{Retrieval alignment.}
For a batch of $B$ matched text--sample pairs we form two $B\times B$
score matrices, differing only in \emph{which} text conditions the
weights,
\begin{equation}
\begin{aligned}
    C_{ij}&=\langle z_i^T,\mu_{\mathcal{M}_j}(z_i^T)\rangle,
    \\[2pt]
    \tilde{C}_{ij}&=\langle z_i^T,\mu_{\mathcal{M}_j}(z_j^T)\rangle ,
\end{aligned}
\label{eq:scores}
\end{equation}
so that in $C$ every candidate is re-aggregated with respect to the
query that scores it, matching inference, whereas in $\tilde{C}$ each
candidate is aggregated once using its own paired text; $\tilde{C}$ is
used by the semantic objective below. Retrieval is trained with
symmetric InfoNCE~\cite{Radford2021LearningTV,chen2020simple} over $S=C/\tau$,
\begin{equation}
    \mathcal{L}_{\mathrm{align}}
    =
    \frac{1}{2}\left[
        \mathrm{InfoNCE}(S)+\mathrm{InfoNCE}(S^{\top})
    \right],
    \label{eq:align}
\end{equation}
where $\mathrm{InfoNCE}$~\cite{Radford2021LearningTV,chen2020simple} is taken
over the batch with matched pairs on the diagonal, $\tau=0.07$, and
label smoothing $0.1$.

\noindent\textbf{Cross-arity consistency.}
Writing $s_{\mathcal{M}}=\langle z^T,\mu_{\mathcal{M}}(z^T)\rangle$ and
$s_{\mathcal{K}}=\langle z^T,\mu_{\mathcal{K}}(z^T)\rangle$ for the
positive-pair scores of the reduced and complete views of a sample, we
tie the two views through
\begin{equation}
    \mathcal{L}_{\mathrm{mask}}
    =
    1-\langle\mu_{\mathcal{M}},\mu_{\mathcal{K}}\rangle
    +
    \left(s_{\mathcal{M}}-\mathrm{sg}[s_{\mathcal{K}}]\right)^2,
    \label{eq:lmask}
\end{equation}
where $\mathrm{sg}[\cdot]$ denotes the stop-gradient operation. The two terms are coupled to the gradient in distinct ways. The score term treats the complete-set score as a fixed reference, ensuring that calibration aligns the reduced view with the complete view, rather than allowing both to converge at an intermediate value that would not occur during testing. The direction term propagates gradients through both centroids, emphasizing that agreement across modalities is an intrinsic property required of the representation, rather than a target imposed by one view on the other. Freezing $\mu_{\mathcal{K}}$ would exempt the complete-modality view from the constraint it is intended to impose.

\noindent\textbf{Graded semantic supervision.}
A frozen sentence encoder defines affinities
$S^*_{ij}=((\cos(e_i,e_j)+1)/2)^{1/\tau^*}$ with $\tau^*=0.5$,
retaining the top $64$ neighbors per sample. Let $\Omega = \{(i,j) : S^*_{ij} > 0\}$ denote the pairs with an observed target. The cache is sparsified, so an absent entry means the affinity is unknown rather than zero, and the calibration term below is therefore restricted to $\Omega$. The semantic objective combines neighborhood matching with direct cosine calibration,
\begin{equation}
\begin{split}
    \mathcal{L}_{\mathrm{sem}}
    =\ &\mathrm{KL}\!\left(P^*\,\middle\|\,\mathrm{softmax}(\tilde{C}/\tau)\right)\\
    &+\mathbb{E}_{(i,j)\in\Omega}
      \left[\bigl(\tilde{C}_{ij}-(2S^*_{ij}-1)\bigr)^2\right],
\end{split}
\label{eq:lsem}
\end{equation}
with $P^*=\mathrm{softmax}(S^*/\tau^*)$, where the rescaling $2S^*-1$ maps affinities onto the cosine range $[-1,1]$.  The KL term does not admit the same restriction: it is normalized over the full row, and renormalizing over a support that varies per row would make the objective incomparable across rows. Absent entries there consequently enter at the softmax's uniform baseline weight. The ranking term thus treats an unknown
affinity as unremarkable, whereas the calibration term abstains from evaluating it. This objective employs $\tilde{C}$, allowing the graded targets to influence the embedding geometry independently of the query-conditioned weighting.

\noindent\textbf{Uniformity.}
Letting $\mu_i=\mu_{\mathcal{M}_i}(z_i^T)$ denote the aggregated
representation of sample $i$, we prevent global collapse by spreading
the $B$ representations of a batch over the sphere~\cite{wang2020understanding},
\begin{equation}
    \mathcal{L}_{\mathrm{unif}}
    =
    \log\frac{1}{B(B-1)}\sum_{i\neq j}\exp\!\left(-2\|\mu_i-\mu_j\|^2\right).
    \label{eq:lunif}
\end{equation}
\noindent\textbf{Full objective.} Following \gram{}~\cite{cicchetti2025gramian}, we retain the data-anchor matching loss $\mathcal{L}_{\mathrm{DAM}}$, which trains the second-stage matching head with hard negatives, at its published weight of $0.1$. The complete training objective is
\begin{equation}
    \mathcal{L}
    =
    \mathcal{L}_{\mathrm{align}}
    +\mathcal{L}_{\mathrm{sem}}
    +\mathcal{L}_{\mathrm{mask}}
    +0.1\,\mathcal{L}_{\mathrm{unif}}
    +0.1\,\mathcal{L}_{\mathrm{DAM}} .
    \label{eq:total}
\end{equation}
We apply the semantic and uniformity terms after a 500-step warm-up. Our method adds no aggregation network. We freeze the pretrained backbone and adapt its query and value projections with rank-8 LoRA~\cite{hu2021lora}
($\alpha = 16$), together with the lightweight projection and matching heads, giving 4.8 million trainable parameters against the full-backbone pretraining
used by the compared methods. Sec.~\ref{sec:supp-parameters} explains why we do not assign a trainable-parameter count to released checkpoints. Algorithm~\ref{alg:supp-train} summarizes one training step.

\noindent\textbf{Inference} The standard two-stage retrieval protocol is retained. \ours first scores the full candidate set using Eq.~\eqref{eq:closedform} and passes the top-$50$ samples to the backbone's cross-encoder for final reranking. For incomplete samples, Eq.~\eqref{eq:qcentroid} is evaluated only over the observed set $\mathcal{M}$. No modality is imputed, and no sample is discarded. The same cosine-based score applies unchanged across modality subsets.

\section{Experiments}
\label{sec:experiments}

\noindent\textbf{Datasets.}
We conduct evaluations on MSR-VTT~\cite{MSRVTT}, DiDeMo~\cite{DIDEMO}, ActivityNet Captions~\cite{ACTIVITYNET}, VATEX~\cite{wang2019vatex}, and AudioCaps~\cite{AUDIOCAPS}, adhering to the dataset splits and modality configurations described in~\cite{cicchetti2025gramian}. MSR-VTT and VATEX include video, audio, and subtitles, while the other datasets comprise video and audio only. For VATEX, we use the 431 downloadable test clips to ensure a consistent gallery
across all methods. Additional dataset statistics are given in Table~\ref{tab:supp-datastats}.

\noindent\textbf{Implementation Details.} The implementation builds upon VAST~\cite{Chen2023VASTAV}, which employs EVA-CLIP. ViT-g/14\footnote{https://huggingface.co/QuanSun/EVA-CLIP}~\cite{Sun2023EVACLIPIT} is utilized for video, BEATs\footnote{github.com/microsoft/unilm}~\cite{beats2023} for audio, and BERT-base\footnote{https://huggingface.co/google-bert/bert-base-uncased}~\cite{devlin2019bert} for both subtitles and text. Whereas baseline methods pretrain the entire VAST model, the proposed approach freezes the backbone and trains rank-8 LoRA adapters~\cite{hu2021lora} ($\alpha{=}16$) on the query and value projections, resulting in 4.8 million trainable parameters. Training is initialized from the VAST pretrained model and proceeds on the same 150,000-clip subset of VAST-27M used by the baselines; 136,674 of these clips remain downloadable. We train for five epochs on four NVIDIA A100 GPUs using AdamW~\cite{loshchilov2017decoupled}, with a learning rate of $2\times10^{-5}$ and a batch size of 128. The query-weighting temperature is set to $\tau_w{=}0.1$, and the contrastive temperature is fixed at $\tau{=}0.07$. Reduced-arity views are sampled by annealing $p_{\mathrm{full}}$ from 1 to 0.5 over the first 2,000 steps. Complete hyperparameter settings are provided in Table~\ref{tab:supp-hparams}.
\begin{table}[t]
\centering
\caption{Zero-shot retrieval on MSR-VTT. $\S$: published numbers (reference only). $\star$: authors' released checkpoint, evaluated on our protocol. $\dagger$: trained from the authors' released code at their recipe. \ours{}: three seeds, $\pm$ sd on R@1.}
\label{tab:msrvtt}
\small
\setlength{\tabcolsep}{4pt}
\resizebox{\columnwidth}{!}{%
\begin{tabular}{llccccc}
\toprule
 & & & \multicolumn{2}{c}{Text $\rightarrow$ Video} & \multicolumn{2}{c}{Video $\rightarrow$ Text} \\
\cmidrule(lr){4-5}\cmidrule(lr){6-7}
Method & Adapter & Mask & R@1 & R@10 & R@1 & R@10 \\
\midrule
\multicolumn{7}{l}{\emph{(a) Foundation models}} \\
% ImageBind$^{\S}$~\cite{Girdhar2023ImageBindOE} & -- & \xmark & -- & -- & -- & -- \\
UMT-L (25M)$^{\S}$ & -- & \xmark & 40.7 & 71.8 & -- & -- \\
LanguageBind$^{\S}$~\cite{Zhu2023LanguageBindEV} & -- & \xmark & 44.8 & 78.7 & 40.9 & 75.7 \\
mPLUG-2$^{\S}$~\cite{Xu2023mPLUG2AM} & -- & \xmark & 47.1 & 79.0 & -- & -- \\
VideoPrism-b$^{\S}$~\cite{Zhao2024VideoPrismAF} & -- & \xmark & 51.4 & -- & 50.2 & -- \\
VAST (27M)$^{\S}$~\cite{Chen2023VASTAV} & full-FT & \xmark & 50.7 & 74.4 & -- & -- \\
\midrule
\multicolumn{7}{l}{\emph{(b) Gramian-volume alignment}} \\
\gram{}$^{\star}$~\cite{cicchetti2025gramian} & full-FT & \xmark & 52.5 & 82.5 & 50.5 & \textbf{81.2} \\
HyperGRAM$^{\dagger}$~\cite{na2026hypergram} & full-FT & \xmark & 54.0 & 82.0 & 51.9 & 81.1 \\
\midrule
\multicolumn{7}{l}{\emph{(c) Leading-eigenvalue alignment}} \\
PMRL$^{\star}$~\cite{liu2026principled} & full-FT & \xmark & 54.3 & 79.5 & \textbf{52.9} & 79.4 \\
\midrule
\multicolumn{7}{l}{\emph{(d) Spherical centroid alignment (ours)}} \\
\textbf{\ours} (ours) & LoRA & \cmark & \textbf{54.6} & \textbf{84.9} & 50.5 & 78.7 \\
\ours, full-FT (same recipe) & full-FT & \cmark & 50.4 & 82.1 & 50.5 & 80.4 \\
\bottomrule
\end{tabular}}
\end{table}
\begin{table*}[t]
\centering
\caption{Zero-shot text-to-video retrieval on the transfer benchmarks. $\S$: numbers as published (reference only; not comparable across environments). $\star$: the authors' released checkpoint evaluated in our environment on our protocol. $\dagger$: trained from the authors' unmodified code at their published recipe (no checkpoint is released). \ours{} is one configuration over three seeds ($\pm$ sd on R@1).}
\label{tab:transfer}
\small
\setlength{\tabcolsep}{4pt}
\resizebox{1.5\columnwidth}{!}{%
\begin{tabular}{llccccccccc}
\toprule
 & & & \multicolumn{2}{c}{DiDeMo} & \multicolumn{2}{c}{ActivityNet} & \multicolumn{2}{c}{VATEX} & \multicolumn{2}{c}{AudioCaps} \\
\cmidrule(lr){4-5}\cmidrule(lr){6-7}\cmidrule(lr){8-9}\cmidrule(lr){10-11}
Method & Adapter & Mask & R@1 & R@10 & R@1 & R@10 & R@1 & R@10 & R@1 & R@10 \\
\midrule
\multicolumn{11}{l}{\emph{(a) Foundation models}} \\
% ImageBind$^{\S}$ & -- & \xmark & -- & -- & -- & -- & -- & -- & -- & -- \\
UMT-L (25M)$^{\S}$ & -- & \xmark & 48.6 & 79.0 & 41.9 & -- & -- & -- & -- & -- \\
LanguageBind$^{\S}$ & -- & \xmark & 39.9 & 74.6 & 41.0 & 80.0 & -- & -- & -- & -- \\
mPLUG-2$^{\S}$ & -- & \xmark & 45.7 & 71.1 & -- & -- & -- & -- & -- & -- \\
VideoPrism-b$^{\S}$ & -- & \xmark & -- & -- & 49.6 & -- & 62.5 & -- & -- & -- \\
VAST (27M)$^{\S}$ & full-FT & \xmark & 49.5 & 76.9 & 51.4 & 83.6 & 82.1 & 96.8 & -- & -- \\
\midrule
\multicolumn{11}{l}{\emph{(b) Gramian-volume alignment}} \\
\gram{}$^{\star}$~\cite{cicchetti2025gramian} & full-FT & \xmark & 50.7 & 76.5 & \textbf{56.3} & 87.0 & 90.0 & \textbf{100.0} & 32.2 & 74.3 \\
\hypergram{}$^{\dagger}$~\cite{na2026hypergram} & full-FT & \xmark & 48.9 & 76.7 & 53.7 & 86.2 & 89.8 & \textbf{100.0} & -- & -- \\
\midrule
\multicolumn{11}{l}{\emph{(c) Leading-eigenvalue alignment}} \\
PMRL$^{\star}$~\cite{liu2026principled} & full-FT & \xmark & \textbf{52.5} & \textbf{79.9} & 54.1 & 85.3 & 89.6 & 98.8 & 34.4 & \textbf{76.3} \\
\midrule
\multicolumn{11}{l}{\emph{(d) Spherical centroid alignment (ours)}} \\
\textbf{\ours} (ours) & LoRA & \cmark & 51.2 & 78.9 & 55.9 & \textbf{87.2} & \textbf{90.6} & 99.6 & \textbf{35.0} & 74.9 \\
\ours, full-FT (same recipe) & full-FT & \cmark & 49.0 & 75.7 & 53.4 & 85.7 & 89.3 & 99.3 & 33.0 & 73.9 \\
\bottomrule
\end{tabular}}
\end{table*}

\noindent\textbf{Baselines.} This study compares \gram{}\footnote{https://github.com/ispamm/GRAM}~\cite{cicchetti2025gramian}, \pmrl{}\footnote{https://github.com/Xiaohao-Liu/PMRL}~\cite{liu2026principled}, and HyperGRAM\footnote{https://github.com/uta-smile/HyperGram}~\cite{na2026hypergram}, which are three geometric aggregation methods built on a shared backbone architecture. We use \gram{} as the primary baseline. Table~\ref{tab:supp-environment} quantifies the spread by scoring the same GRAM checkpoint in three environments. Repeated runs vary by at most 0.2 R@1, which we treat as evaluation noise (Section~\ref{sec:supp-repro}).
\subsection{Zero-Shot Retrieval}
\label{sec:zeroshot}
Tables~\ref{tab:msrvtt} and~\ref{tab:transfer} report zero-shot retrieval results across the five benchmarks under the evaluation protocol of~\cite{cicchetti2025gramian}. \ours{} achieves the highest text-to-video R@1 on MSR-VTT ($54.6$), outperforming \pmrl{} ($54.3$) and \gram{} ($52.5$), as well as achieving 90.6 on VATEX and 35.0 on AudioCaps. On ActivityNet, \ours{} attains $55.9\pm0.2$, which is comparable to \gram{}'s $56.3$. On DiDeMo, \ours{} is $1.3$ R@1 below \pmrl{}. The relative ranking varies by adaptation strategy and retrieval direction.
\noindent\textbf{Low-rank and full-model adaptation.} Full fine-tuning control employs the same training procedure as the proposed method, but updates the entire backbone instead of only the 4.8 million adapter parameters. In this configuration, text-to-video R@1 decreases by 4.2 on MSR-VTT, 2.2 on DiDeMo, 2.5 on ActivityNet, 1.3 on VATEX, and 2.0 on AudioCaps, with an average reduction of 2.4 R@1 across the five benchmarks. These findings indicate that low-rank adaptation is not merely a computational compromise. In the text-to-video setting, it also aligns more closely with the proposed training objective. However, the ranking reverses for video-to-text R@1, where full fine-tuning is comparable to or outperforms low-rank adaptation on all five benchmarks. Thus, the relative advantage of each adaptation strategy depends on the retrieval direction.

\noindent\textbf{Reverse-direction retrieval.}
This direction dependence is also evident in comparisons with baseline methods. For video-to-text R@1, \ours{} remains competitive but does not consistently match the strongest baseline. For instance, on MSR-VTT, \ours{} achieves $50.5$, whereas PMRL attains $52.9$. This asymmetry aligns with the objective's design, which conditions modality aggregation on the text query and thus more directly supports text-to-video retrieval. Reporting the reverse direction clarifies the scope of the observed gains without assuming symmetric transfer.

\subsection{Aggregation Gain}
\label{sec:gain}

Table~\ref{tab:gain} presents the aggregation gain, defined as the difference between the joint score of a method and its strongest single-modality pathway. A positive value indicates that aggregation improves performance over each constituent pathway. Both scores are computed from the same checkpoint and embeddings, so this within-method comparison is unaffected by the environment-dependent offsets documented in Table~\ref{tab:supp-environment}. Constituent scores are given in Table~\ref{tab:supp-gain-components}.
\begin{table}[t]
\centering
\caption{\textbf{Retrieval under missing modalities.}
Absolute text$\rightarrow$video R@1 using each method's own aggregation score when one modality is removed from a fraction \(r\) of gallery clips. Masks are deterministic and identical across methods; \(r{=}0\) reproduces the main protocol. Figure~\ref{fig:gain_vs_mask} reports the complementary gain over each method's unimodal pathway. \(\star\): authors' released checkpoint. For \textsc{Gram} a missing modality is made an orthonormal axis rather than
zero-filled, contributing a factor of one, so the determinant reduces to the sub-Gramian over the modalities each clip retains; zero-filling would instead
force $\det G {=} 0$ for every masked clip and collapse the ranking among them
(Sec.~\ref{sec:maskedvol}). Two-stage results under the same masks are analyzed in Sec.~\ref{sec:supp-two-stage}. PMRL is excluded because its released scoring does not reproduce the \(r{=}0\) result through the masking harness, while \hypergram{} provides no missing-modality evaluation path.}
\label{tab:missing}
\small
\setlength{\tabcolsep}{4.5pt}
\resizebox{0.9\columnwidth}{!}{%
\begin{tabular}{llccccc}
\toprule
 & Method & 0\% & 25\% & 50\% & 75\% & 90\% \\
\midrule
MSR-VTT & \textbf{\ours} (ours) & \textbf{45.2} & \textbf{42.1} & \textbf{39.2} & \textbf{36.3} & \textbf{34.5} \\
 & \gram{}$^{\star}$ & 38.7 & 36.6 & 32.4 & 29.9 & 27.9 \\
\rowcolor{gray!10} & \emph{margin} & +6.5 & +5.5 & +6.8 & +6.4 & +6.6 \\
\midrule
DiDeMo & \textbf{\ours} (ours) & \textbf{34.3} & \textbf{31.1} & \textbf{27.5} & \textbf{24.0} & \textbf{22.5} \\
 & \gram{}$^{\star}$ & 28.2 & 24.8 & 22.0 & 19.7 & 18.1 \\
\rowcolor{gray!10} & \emph{margin} & +6.1 & +6.3 & +5.5 & +4.3 & +4.4 \\
\midrule
ActivityNet & \textbf{\ours} (ours) & \textbf{34.4} & \textbf{30.3} & \textbf{27.4} & \textbf{24.7} & \textbf{23.0} \\
 & \gram{}$^{\star}$ & 31.0 & 27.1 & 23.6 & 20.8 & 19.5 \\
\rowcolor{gray!10} & \emph{margin} & +3.4 & +3.2 & +3.8 & +3.9 & +3.5 \\
\midrule
VATEX & \textbf{\ours} (ours) & \textbf{81.7} & \textbf{78.7} & \textbf{71.5} & \textbf{64.0} & \textbf{61.0} \\
 & \gram{}$^{\star}$ & 75.6 & 69.8 & 61.7 & 53.1 & 51.5 \\
\rowcolor{gray!10} & \emph{margin} & +6.1 & +8.9 & +9.8 & +10.9 & +9.5 \\
\midrule
AudioCaps & \textbf{\ours} (ours) & \textbf{27.1} & \textbf{23.7} & \textbf{20.5} & \textbf{16.8} & \textbf{13.9} \\
 & \gram{}$^{\star}$ & 22.9 & 20.3 & 17.0 & 12.2 & 10.1 \\
\rowcolor{gray!10} & \emph{margin} & +4.2 & +3.4 & +3.5 & +4.6 & +3.8 \\
\bottomrule
\end{tabular}}
\end{table}
For geometric aggregators, the joint score is lower than the strongest single-modality pathway in 12 out of 14 measurable method-by-benchmark combinations (one-sided sign test,
$p{=}0.0065$). \gram{} demonstrates negative gains across all five benchmarks (ranging from $-1.9$ to $-6.1$), PMRL on four out of five (down to $-27.6$), and HyperGRAM on three out of four. The exceptions are PMRL on AudioCaps ($+1.4$) and HyperGRAM on VATEX ($+0.0$), where the constituent pathways are similarly strong. These results suggest that symmetric aggregation is less sensitive to pathway imbalance when the modalities provide comparably informative signals, although dataset-specific factors may also contribute. In comparison, \ours{} achieves positive gains on four out of five benchmarks, including $+4.0$ on MSR-VTT, and a negative gain of $-1.1$ on DiDeMo.
\begin{table*}[t]
\centering
\caption{\textbf{Aggregation gain:} joint multimodal T$\rightarrow$V R@1 minus the strongest single-modality R@1 from the same model. Positive values are bold; negative values indicate reduced performance. $\star$: authors' released checkpoint; $\dagger$: trained from the authors' code. The uniform-weight variant removes query conditioning while retaining the centroid and model trunk; Table~\ref{tab:loss_ablation} examines masked-view training. Baseline gains are computed from each model's embeddings using its published scoring function in a shared environment. Constituent scores are given in Tab.~\ref{tab:supp-gain-components}.
}
\label{tab:gain}
\small
\setlength{\tabcolsep}{5pt}
\resizebox{1.3\columnwidth}{!}{%
\begin{tabular}{lccccc}
\toprule
Method & MSR-VTT & DiDeMo & ActivityNet & VATEX & AudioCaps \\
\midrule
\gram{}$^{\star}$~\cite{cicchetti2025gramian} & -3.4 & -3.8 & -5.8 & -1.9 & -6.1 \\
\pmrl{}$^{\star}$~\cite{liu2026principled} & -12.2 & -9.6 & -9.8 & -27.6 & \textbf{+1.4} \\
\hypergram{}$^{\dagger}$~\cite{na2026hypergram} & -3.4 & -0.1 & -2.0 & +0.0 & -- \\
\midrule
\ours, uniform weights & -7.0 & -7.5 & -4.0 & -11.2 & \textbf{+0.2} \\
\textbf{\ours, query-weighted (ours)} & \textbf{+4.0} & -1.1 & \textbf{+0.3} & \textbf{+0.5} & \textbf{+1.4} \\
\bottomrule
\end{tabular}}
\end{table*}

\noindent\textbf{Contribution of Query Conditioning.} To assess the impact of query conditioning, the uniform-weight control substitutes query-dependent weights with a simple mean across modality embeddings, while maintaining the same trunk and training procedure. This approach results in aggregation gains of $-7.0$, $-7.5$, $-4.0$, and $-11.2$ on MSR-VTT, DiDeMo, ActivityNet, and VATEX, respectively. The observed decline persists even without a determinant, hyperbolic volume, or eigenspectrum, suggesting that uniform modality weighting, rather than any specific higher-order construction, is responsible. The modality-subset analysis in Table~\ref{tab:supp-subset-ladder} reveals a similar pattern on MSR-VTT: the score of SCALAR increases from $41.2$ to $42.1$ and then $45.2$ as modalities are added, whereas the corresponding volume-based scores decrease from $42.1$ to $39.7$ and then $38.7$. Collectively, these comparisons demonstrate that query conditioning substantially contributes to aggregation gain.
\subsection{Robustness to Test-Time Missing Modalities}
\label{sec:missing}

In practical scenarios, galleries may include incomplete modality sets. The available volume-based pipelines are not explicitly designed for this context and exclude clips with missing modalities during data loading. Therefore, this study evaluates the behavior of the respective scoring functions when a modality is unavailable at test time.

\noindent\textbf{Protocol.} A fraction $r \in \{0, 25, 50, 75, 90\}\%$ of gallery clips is assigned a masked modality. Masks are deterministically generated for each clip by hashing a fixed seed with the clip identifier, nested across rates, and shared among all methods. A masked modality is written as zeros in the common trunk, matching what a loader produces for an absent stream; presence is then recovered from the embedding norm before any score is formed, so the zeros are a transport convention and never reach an aggregation function as a vector. Each method applies its aggregation function to the modalities retained for each clip. The masked centroid is defined for any non-empty subset of modalities. In \gram{}, the absent axis is made orthonormal rather than zero-filled: This approach ensures that the absent axis contributes a factor of exactly one to the determinant, which therefore reduces to the sub-Gramian over the modalities the clip retains—the method's established lower-arity formula, already implemented for arities 2 to 4 across datasets. In contrast, zero-filling would introduce both a zero row and column in the Gram matrix, forcing $\det G = 0$ for every masked clip and thereby collapsing the ranking among them. This approach is not adopted; instead, the baseline is given a treatment that its release does not support, which discards incomplete clips outright (Sec.~\ref{sec:maskedvol}). The $r{=}0$ setting is byte-identical to the standard protocol and reproduces the main results. PMRL's released weights can be loaded into the evaluation harness; however, its score at $r{=}0$ does not match the repository result ($49.8$ versus $54.3$ R@1 on MSR-VTT). Therefore, we omit masked results for PMRL, as they would not provide a reliable comparison. HyperGRAM's release does not include a missing-modality evaluation path.

Each method employs a checkpoint-specific cross-encoder for reranking, with rerankers trained exclusively on complete modality sets. Under test-time masking, the two-stage result consequently reflects both the aggregation rule and the reranker's response to inputs outside its training distribution.
This overlap makes the first-stage effect difficult to isolate. At $r{=}90\%$, the two-stage R@1 values closely match the corresponding video-only cosine scores: \ours{} achieves $33.2$ compared to $32.1$, and \gram{} achieves $32.9$ compared to $32.8$ (Table~\ref{tab:supp-missing-two-stage}). The performance gap between methods also decreases from $2.3$ points to within the range of evaluation variation on most benchmarks. Consequently, Table~\ref{tab:missing} reports each method's first-stage score, which is the component directly influenced by the aggregation intervention. Complete two-stage results for every masking rate are given in Section~\ref{sec:supp-two-stage}. The convergence of the two-stage scores further indicates that robustness to missing modalities should be evaluated at both the representation and reranking stages.

\begin{figure*}[h]
\centering
\includegraphics[width=0.75\linewidth]{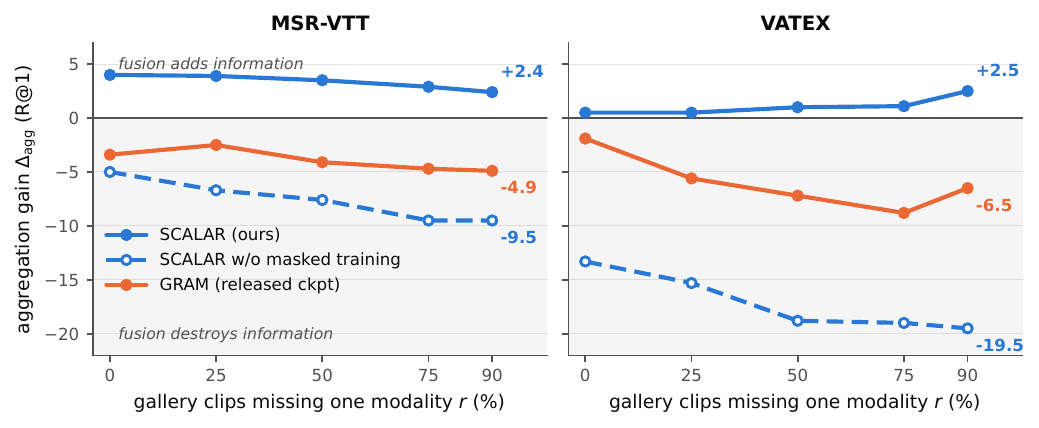}
\caption{\textbf{Aggregation gain under missing modalities.} Gain is each method's masked score minus its own strongest unimodal score under the same masks; Table~\ref{tab:missing} reports absolute accuracy. \ours{} remains positive at all evaluated rates, whereas \gram{} declines as masking increases. \ours{} without masked-view training shows a similar decline, suggesting that both query conditioning and masked-view training help preserve aggregation benefits (Tables~\ref{tab:gain} and~\ref{tab:loss_ablation}).}
\label{fig:gain_vs_mask}
\end{figure*}
\noindent\textbf{Results.} At every masking rate on every benchmark, \ours obtains the highest first-stage
R@1 among the methods evaluated (exact one-sided sign test over the individual cells, $p=3\times10^{-8}$; treating each benchmark as a single unit rather than each cell, $p=0.031$). The observed margin ranges from $3.2$ to 10.9 R@1. It remains relatively stable on MSR-VTT and ActivityNet, increases on VATEX ($+6.1$ to $+10.9$ at 75\% masking), and decreases on DiDeMo ($+6.1$ to $+4.4$) and AudioCaps ($+4.2$ to $+3.8$). Figure~\ref{fig:gain_vs_mask} provides the corresponding within-method view. Relative to its own strongest single pathway, \ours{} retains positive aggregation gain across all masking rates on MSR-VTT ($+4.0$ to $+2.4$) and VATEX ($+0.5$ to $+2.5$), whereas GRAM's gain on VATEX falls from $-1.9$ to $-8.8$ at 75\% masking. On AudioCaps, the gain of \ours{} becomes negative at high masking rates, suggesting that the remaining modalities
do not fully compensate when the principal audio signal is absent.
\begin{table*}[t]
\centering
\caption{Objective ablation: the reported configuration with one component removed, nothing retuned. The reference row shows absolute values; ablation rows show the \emph{change} caused by the removal. $\bar{\Delta}$: mean two-stage text$\rightarrow$video R@1. $\bar{\Delta}_{\mathrm{agg}}$: mean aggregation gain (Table~\ref{tab:gain}); $\bar{\Delta}_{\mathrm{agg}}^{90}$: the same under 90\% test-time masking. Regularizers are not swept ('--').}
\label{tab:loss_ablation}
\small
\setlength{\tabcolsep}{4pt}
\resizebox{1.5\columnwidth}{!}{%
\begin{tabular}{lccccc ccc}
\toprule
Objective & MSR-VTT & DiDeMo & ActivityNet & VATEX & AudioCaps & $\bar{\Delta}$ & $\bar{\Delta}_{\mathrm{agg}}$ & $\bar{\Delta}_{\mathrm{agg}}^{90}$ \\
\midrule
\multicolumn{9}{l}{\emph{(a) core (not ablatable: $\mathcal{L}_{\mathrm{align}}$ is the objective)}} \\
full objective (T9) & 54.8 & 51.5 & 55.8 & 90.5 & 35.2 & -- & +1.0 & -0.5 \\
\midrule
\multicolumn{9}{l}{\emph{(b) mechanism: masked-view training}} \\
\quad w/o masked training views ($p_{\mathrm{full}}{=}1$) & 53.8 & 50.4 & 54.7 & 90.5 & 34.5 & -0.8 & -6.5 & -7.1 \\
\quad w/o $\mathcal{L}_{\mathrm{mask}}$ ($\beta{=}0$) & 54.8 & 51.0 & 55.8 & 90.3 & 35.4 & -0.1 & -1.5 & +0.0 \\
\midrule
\multicolumn{9}{l}{\emph{(c) regularizers}} \\
\quad w/o $\mathcal{L}_{\mathrm{sem}}$ ($\alpha{=}0$) & 54.2 & 50.2 & 56.3 & 91.2 & 35.8 & -0.0 & -0.9 & -- \\
\quad w/o $\mathcal{L}_{\mathrm{unif}}$ ($\lambda{=}0$) & 54.2 & 51.2 & 56.0 & 90.5 & 35.1 & -0.2 & -0.4 & -- \\
\bottomrule
\end{tabular}}
\end{table*}

\noindent\textbf{Effect of masked-view training.}
Omitting masked views during training degrades the text--audio pathway by 17 to 39\% relative across the five benchmarks while leaving the video pathway essentially unchanged. At $r{=}0$, the aggregation gain falls from $+4.0$ to $-5.0$ on MSR-VTT and from $+0.5$ to $-13.3$ on VATEX; at $r{=}90\%$ the corresponding values are $-9.5$ and $-19.5$ (Fig.~\ref{fig:gain_vs_mask}, open markers). This difference occurs with minimal change to the complete-modality two-stage score (Tab.~\ref{tab:loss_ablation}). This comparison demonstrates that query-conditioned weighting and masked-view training provide complementary benefits: the former
adjusts modality relevance for each query, while the latter exposes the model to variations in modality availability.
\subsection{Objective Ablation}
\label{sec:ablation}

Table~\ref{tab:loss_ablation} removes each objective component from the reported
configuration without further retuning. At the two-stage metric, the resulting differences remain within seed variation, consistent with the compression introduced by reranking in \S\ref{sec:missing}. We therefore also report mean
aggregation gain and its masked counterpart, which more directly reflect the components targeted by these losses.

Masked-view training produces the largest ablation effect: removing it decreases aggregation gain by $6.5$ points, or $7.1$ points under masking, while altering the two-stage score by only $-0.8$ and maintaining stability in the video pathway. Removing $\mathcal{L}_\text{mask}$ costs 1.5 points of aggregation gain while
changing the two-stage score by only 0.1. Excluding $\mathcal{L}_{\mathrm{sem}}$ or $\mathcal{L}_{\mathrm{unif}}$ reduces the gain by $0.9$ and $0.4$ points, respectively, without impacting parameter count or inference cost. These findings indicate that robustness primarily results from reduced-arity training and cross-arity agreement, with regularizers contributing to a lesser extent. The sign tests behind these counts are given in Section~\ref{sec:supp-statistics}.

\subsection{Analysis of Remaining Performance Gaps}
\label{sec:deficits}

In the final two-stage evaluation, \ours{} lags behind PMRL by $1.3$ R@1 on DiDeMo and behind \gram{} by $0.4$ on ActivityNet, with the latter difference approaching the margin of evaluation variation. These performance gaps arise during method-specific reranking: prior to
reranking, SCALAR outperforms GRAM on all five benchmarks by 3.4 to 6.5 R@1 (Table~\ref{tab:supp-gain-components}). On DiDeMo, reranking increases PMRL's score by $23.8$ points, compared to a $ 16.9$-point improvement for \ours{}; on ActivityNet, \gram{} gains $25.3$ points while \ours{} gains $21.5$. These results indicate that reranker compatibility partially accounts for the remaining performance differences, which aligns with the narrowing observed when modalities are missing. In VGGSound audio-visual classification, \ours{} shows lower transfer performance than GRAM (34.9$\pm$0.3 vs. 40.5 Acc@1). This performance gap is primarily due to the video pathway (31.3 vs. 35.0), rather than the aggregation method. Notably, \ours{}'s multimodal score continues to exceed its strongest unimodal pathway by
3.6 points. The GRAM result aligns with its published value within 0.1 Acc@1, thereby validating the evaluation protocol. Section~\ref{sec:supp-vggsound} details the decomposition, and Section~\ref{sec:supp-latent} presents the latent-space comparison.

% The supplement presents an evaluation of transfer to VGGSound audio-visual classification. \ours{} achieves $34.9{\pm}0.3$ Acc@1, compared with $40.5$ for \gram{}. Most of this performance gap is attributable to the video pathway ($31.3$ vs. $35.0$), while \ours{} shows a slightly stronger audio pathway ($27.0$ vs. $26.3$). Notably, \ours{} 's multimodal score exceeds its strongest unimodal pathway by $3.6$ points, indicating positive within-checkpoint aggregation despite weaker transfer performance on this task. The \gram{} result also matches the published value within $0.1$ Acc@1 ($40.5$ vs. $40.6$), supporting the consistency of the evaluation protocol. A latent-space comparison against the shared initialization is given in Section~\ref{sec:supp-latent}, and a transfer result on VGGSound, where \ours does not lead, in Section~\ref{sec:supp-vggsound}.

\section{Conclusion}
In this paper, we re-examine geometric multimodal aggregation against a
straightforward criterion: whether aggregation improves on the strongest
individual modality within the same representation. Under a unified evaluation
protocol, symmetric geometric aggregators frequently fail this criterion,
motivating \ours{}, which applies a query-conditioned weighting to each
modality's contribution before computing a normalized spherical centroid.
Across five retrieval benchmarks, \ours{} achieves positive aggregation gain on
four while updating 4.8 million parameters, and outperforms GRAM at the
representation stage across all masking rates on every benchmark. Ablation
studies identify query-conditioned weighting and masked reduced-arity training
as the primary contributors. However, frozen rerankers can eliminate these
first-stage gains when modalities are absent.
% This work revisited geometric multimodal aggregation using a straightforward criterion: whether aggregation improves upon the strongest available modality within the same representation. Under a unified evaluation protocol, symmetric geometric aggregators frequently do not meet this criterion, which motivated the development of \ours. \ours conditions each modality's contribution on the query and forms a normalized spherical centroid. Across five retrieval benchmarks, \ours demonstrates positive aggregation gain on four, achieving a +4.0 R@1 improvement while updating only 4.8 million parameters. During test-time modality masking, \ours{} outperforms GRAM in all 25 first-stage settings. Ablation studies identify query-conditioned weighting and masked reduced-arity training as the primary contributors to these results. 

% The analysis further indicates that frozen rerankers can diminish first-stage gains when modalities are absent. Collectively, these findings support query-dependent relevance and subset-aware training as effective principles for robust multimodal retrieval.

% \input{sec/2_formatting}

{
    \small
    \bibliographystyle{ieeenat_fullname}
    \bibliography{main}
}
% {
%     \small
%     \bibliographystyle{ieeenat_fullname}
%     \bibliography{main}
% }

\clearpage
\appendix
\setcounter{table}{0}\renewcommand{\thetable}{S\arabic{table}}
\setcounter{figure}{0}\renewcommand{\thefigure}{S\arabic{figure}}
\setcounter{equation}{0}\renewcommand{\theequation}{S\arabic{equation}}
\setcounter{algorithm}{0}\renewcommand{\thealgorithm}{S\arabic{algorithm}}
% Supplementary material included from main.tex.
\section{Use of Large Language Models}
\label{sec:supp-llm}
% -----------------------------------------------------------------------------

Large language models were used to edit author-written text for grammar,
clarity, terminology, and length. The technical content, experimental design,
reported results, and interpretation were determined by the authors. The
authors checked the revised text against the experimental records and take
responsibility for the complete paper and supplementary material.

% -----------------------------------------------------------------------------
\section{Reproducibility}
\label{sec:supp-repro}
% -----------------------------------------------------------------------------
We will release the training and evaluation code, configuration files, scripts for generating tables and figures, and adapter weights corresponding to the reported configuration. All reported \ours{} results employ a single configuration across the five retrieval benchmarks; no benchmark-specific learning rate, temperature, or schedule is used. Seeds 50, 51, and 52 differ solely in their random seed. The full-fine-
tuning control follows the same training procedure but updates the full model instead of the low-rank adapters. The uniform-weight control disables query conditioning while retaining the same model and training procedure. Each objective ablation in Table~\ref{tab:loss_ablation} removes one component without retuning the remaining hyperparameters.

We evaluate GRAM and PMRL using checkpoints released by their respective authors. As HyperGRAM does not provide a checkpoint, it is trained from the authors' released code following their published recipe. All evaluated methods utilize the VAST foundation model as the initial backbone and are assessed within the same evaluation environment and retrieval protocol. Published numbers are retained solely as reference values in the main paper. Code and checkpoints will be released publicly.

\paragraph{Deterministic missing-modality masks.}
For a clip identifier $c$ and mask seed $\sigma$, we compute
$H=\operatorname{MD5}(\sigma\,\|\,c)$. The first 32-bit block of $H$ defines a
uniform value $u_c\in[0,1)$, and a clip is masked at rate $r$ when $u_c<r$.
The next 32-bit block selects the modality index modulo the number of available
modalities. Because the selected index is independent of $r$, the masked sets
are nested: every clip masked at $r$ remains masked at any $r'>r$. The same
clip-level masks are used for every method, making each comparison paired.

\paragraph{Evaluation variation.}
Repeated evaluation of identical checkpoints over 15 cells produced an
observed spread of at most $0.2$ R@1. We therefore interpret differences of
this magnitude cautiously. Training is not bitwise deterministic because it
uses mixed precision and four-way data-parallel reduction; consequently,
\ours{} results in the main comparison are reported as the mean and standard
deviation over three seeds.

\paragraph{Compute.}
One \ours{} pretraining run takes approximately 11 hours on four NVIDIA A100
64\,GB GPUs. Including the three reported seeds, controls, objective
ablations, missing-modality evaluations, and baseline runs, the complete
experimental study used approximately 900 GPU-hours.

% -----------------------------------------------------------------------------
\section{Algorithms}
\label{sec:supp-algorithms}
% -----------------------------------------------------------------------------

Algorithm~\ref{alg:supp-score} gives the first-stage score for one
query--candidate pair. The query--modality similarities require
$O(|M|d)$ operations. Once those similarities are available, the normalized
weighted sum can be evaluated using the candidate's cached $|M|\times|M|$
Gram matrix with $O(|M|^2)$ scalar operations; no $d$-dimensional centroid must
be stored for every query--candidate pair. A missing modality is excluded from
$M$, and no embedding is imputed.

\begin{algorithm}[t]
\caption{\ours{} score for one query--candidate pair}
\label{alg:supp-score}
\small
\begin{enumerate}[leftmargin=2.1em,label=\arabic*:,itemsep=1pt,topsep=2pt]
  \item \textbf{Input:} query $z^T\in\mathbb{S}^{d-1}$; observed candidate
  embeddings $\{z_m\}_{m\in M}$, where
  $\emptyset\ne M\subseteq\{V,A,S\}$; temperature $\tau_w$.
  \item Compute query agreement
  $a_m\leftarrow\langle z^T,z_m\rangle$ for each $m\in M$.
  \item Normalize the query-conditioned weights over the observed set:
  \[
    w_m\leftarrow
    \frac{\exp(a_m/\tau_w)}{\sum_{n\in M}\exp(a_n/\tau_w)}.
  \]
  \item Retrieve or compute the candidate Gram matrix
  $G_{mn}\leftarrow\langle z_m,z_n\rangle$.
  \item Return
  \[
    s\leftarrow
    \frac{\sum_{m\in M}w_m a_m}
    {\sqrt{\sum_{m,n\in M}w_mw_nG_{mn}}}
    =\left\langle z^T,\mu_M(z^T)\right\rangle.
  \]
\end{enumerate}
\end{algorithm}

Algorithm~\ref{alg:supp-train} summarizes one training step. Masking is applied
after encoding, so the complete and reduced-arity views share a single encoder
forward pass. The reduced view requires only an additional aggregation.

\begin{algorithm}[t]
\caption{One \ours{} training step}
\label{alg:supp-train}
\small
\begin{enumerate}[leftmargin=2.1em,label=\arabic*:,itemsep=1pt,topsep=2pt]
  \item \textbf{Input:} a batch of $B$ matched text--clip pairs, training step
  $t$, and each clip's complete observed set $\mathcal{K}_i$.
  \item Encode each available modality once and $\ell_2$-normalize the
  embeddings.
  \item Set
  $p_{\mathrm{full}}\leftarrow\max(0.5,1-0.5t/2000)$.
  \item For each clip $i$, use $\mathcal{M}_i=\mathcal{K}_i$ with probability
  $p_{\mathrm{full}}$. Otherwise, if $|\mathcal{K}_i|>2$, sample
  $m^\dagger\sim\operatorname{Unif}(\mathcal{K}_i)$ and set
  $\mathcal{M}_i=\mathcal{K}_i\setminus\{m^\dagger\}$. Retain the complete set
  when only two modalities are available.
  \item Form the complete and reduced query-weighted centroids and the two
  score matrices
  \[
    C_{ij}=\left\langle z_i^T,\mu_{\mathcal{M}_j}(z_i^T)\right\rangle,
    \qquad
    \widetilde C_{ij}=
    \left\langle z_i^T,\mu_{\mathcal{M}_j}(z_j^T)\right\rangle.
  \]
  \item Compute symmetric contrastive alignment from $C$:
  \[
    \mathcal{L}_{\mathrm{align}}=
    \tfrac12\!\left[
      \operatorname{InfoNCE}(C/\tau)+
      \operatorname{InfoNCE}(C^\top/\tau)
    \right].
  \]
  \item Average the cross-arity consistency loss over the batch:
  \[
    \mathcal{L}_{\mathrm{mask}}=
    \frac{1}{B}\sum_{i=1}^{B}
    \left[
      1-\left\langle\mu_{\mathcal{M}_i},\mu_{\mathcal{K}_i}\right\rangle
      +\left(s_{\mathcal{M}_i}-
      \operatorname{sg}[s_{\mathcal{K}_i}]\right)^2
    \right].
  \]
  \item After the 500-step warm-up, compute
  $\mathcal{L}_{\mathrm{sem}}$ from $\widetilde C$ and
  $\mathcal{L}_{\mathrm{unif}}$ using Eqs.~(8) and (9) of the main paper;
  before that point, set both terms to zero.
  \item Compute the hard-negative data-anchor matching loss
  $\mathcal{L}_{\mathrm{DAM}}$, drawing negatives using $C$.
  \item Return
  \[
    \mathcal{L}=\mathcal{L}_{\mathrm{align}}+
    \mathcal{L}_{\mathrm{sem}}+\mathcal{L}_{\mathrm{mask}}+
    0.1\mathcal{L}_{\mathrm{unif}}+0.1\mathcal{L}_{\mathrm{DAM}}.
  \]
\end{enumerate}
\end{algorithm}

The matrices $C$ and $\widetilde C$ differ only in the text embedding used to
condition the candidate weights. In $C$, each candidate is re-aggregated for
the query that scores it, matching inference. In $\widetilde C$, each
candidate is aggregated using its paired text, and the resulting matrix is
used only by the graded semantic objective. In
$\mathcal{L}_{\mathrm{mask}}$, the stop-gradient is applied to the complete-set
score $s_{\mathcal{K}}$, which serves as the reference for the reduced view.

% -----------------------------------------------------------------------------
\section{Implementation Details}
\label{sec:supp-implementation}
% -----------------------------------------------------------------------------

Table~\ref{tab:supp-hparams} lists the reported hyperparameters. The same
configuration is used for all five retrieval benchmarks.

\begin{table}[t]
\centering
\caption{Hyperparameters for the reported \ours{} configuration. The three
seeds differ only in the seed value.}
\label{tab:supp-hparams}
\scriptsize
\setlength{\tabcolsep}{3.5pt}
\renewcommand{\arraystretch}{1.03}
\begin{tabular}{@{}ll@{}}
\toprule
\multicolumn{2}{@{}l}{\emph{Frozen backbone}} \\
Initialization & VAST foundation checkpoint \\
Vision encoder & EVA-CLIP ViT-g/14 \\
Audio encoder & BEATs \\
Text encoder & BERT-base (subtitle and query) \\
Projection dimension $d$ & 512 \\
\midrule
\multicolumn{2}{@{}l}{\emph{Adaptation}} \\
LoRA rank / $\alpha$ / dropout & 8 / 16 / 0.0 \\
LoRA target modules & query and value projections \\
Other trainable modules & projection and matching heads \\
Trainable parameters & 4{,}763{,}650 \\
\midrule
\multicolumn{2}{@{}l}{\emph{Optimization}} \\
Optimizer & AdamW, $\beta=(0.9,0.98)$ \\
Learning rate / weight decay & $2\times10^{-5}$ / 0.01 \\
Gradient-norm clip & 2.0 \\
Schedule & linear; warm-up ratio 0.1 \\
Batch size & 128 \\
Epochs / logged steps & 5 / 5{,}330 \\
Precision / parallelism & fp16 / DDP over four GPUs \\
Seeds & 50, 51, 52 \\
\midrule
\multicolumn{2}{@{}l}{\emph{Aggregation and losses}} \\
Query-weight temperature $\tau_w$ & 0.1 \\
Contrastive temperature $\tau$ & 0.07 (fixed) \\
Label smoothing & 0.1 \\
$\mathcal{L}_{\mathrm{sem}}$ weight $\alpha$ & 1.0 \\
$\mathcal{L}_{\mathrm{mask}}$ weight $\beta$ & 1.0 \\
$\mathcal{L}_{\mathrm{unif}}$ weight $\lambda$ & 0.1 \\
$\mathcal{L}_{\mathrm{unif}}$ exponent scale & 2.0 \\
$\mathcal{L}_{\mathrm{DAM}}$ weight & 0.1 \\
Semantic temperature $\tau^*$ & 0.5 \\
Semantic neighbors retained & top 64 \\
Calibration-term weight & 1.0 for observed pairs \\
Loss warm-up & 500 steps \\
\midrule
\multicolumn{2}{@{}l}{\emph{Reduced-arity views}} \\
$p_{\mathrm{full}}$ schedule & $1.0\rightarrow0.5$, linear \\
Annealing duration & 2{,}000 steps \\
Dropped modalities & one, uniform over $\mathcal{K}$ \\
Minimum retained modalities & 2 \\
\midrule
\multicolumn{2}{@{}l}{\emph{Inference}} \\
Sampled frames & 8 \\
Reranking depth & top 50 \\
Evaluation batch size & 64 \\
Test-time mask rates $r$ & 0, 25, 50, 75, 90\% \\
Mask seed & 0 \\
\bottomrule
\end{tabular}
\end{table}

\paragraph{Fixed contrastive temperature.}
We fix $\tau$ at 0.07. The calibration component of
$\mathcal{L}_{\mathrm{sem}}$ supervises the absolute cosine scale, whereas the
softmax component can absorb changes in scale through a learnable temperature.
Fixing $\tau$ keeps the calibrated cosine target and the contrastive scale
separate. The configuration validator therefore rejects the combination of
cosine calibration and a learnable contrastive temperature.

\paragraph{Score matrices used by the objectives.}
The semantic objective uses $\widetilde C$, rather than $C$. This allows the
graded targets to act on the embedding geometry while the retrieval objective
uses the query-conditioned matrix that matches inference.

% -----------------------------------------------------------------------------
\section{Dataset Statistics}
\label{sec:supp-data}
% -----------------------------------------------------------------------------

\begin{table}[t]
\centering
\caption{Dataset statistics for the evaluated protocol. $V$, $A$, and $S$
denote video, audio, and subtitles. VATEX is evaluated on the 431 test clips
that remain downloadable; its absolute recalls are therefore compared only
within the common gallery used in this work.}
\label{tab:supp-datastats}
\scriptsize
\setlength{\tabcolsep}{3pt}
\begin{tabular}{@{}lccccc@{}}
\toprule
Dataset & Modalities & Train & Val & Test & Frames \\
\midrule
MSR-VTT & $V,A,S$ & 9{,}000 & -- & 1{,}000 & 8 \\
DiDeMo & $V,A$ & 8{,}394 & 1{,}065 & 1{,}003 & 8 \\
ActivityNet & $V,A$ & 10{,}009 & -- & 4{,}917 & 8 \\
VATEX & $V,A,S$ & 14{,}060 & -- & 431 & 8 \\
AudioCaps & $V,A$ & -- & -- & 700 & 8 \\
\midrule
VGGSound-5K & $V,A$ & -- & -- & 5{,}000 & 8 \\
\bottomrule
\end{tabular}
\end{table}

\paragraph{Continued-pretraining corpus.}
Our \ours{} runs, controls, ablations, and HyperGRAM retraining use the same
nominal 150k-clip subset of VAST-27M. At the time of our experiments, 136{,}674
of these clips remained downloadable and formed the effective corpus for the
runs we performed. Training uses five epochs; the resulting logs contain 5{,}330
optimizer steps. The released GRAM and PMRL checkpoints are evaluated as
provided, so the exact set of clips available when those checkpoints were
originally trained cannot be reconstructed from the artifacts alone.

\paragraph{Modality availability.}
The retrieval benchmarks cover two observed arities: MSR-VTT and VATEX use
video, audio, and subtitles, whereas DiDeMo, ActivityNet, and AudioCaps use
video and audio. The same \ours{} configuration is therefore evaluated at
$|M|=2$ and $|M|=3$. The test-time masking experiment additionally produces
mixed-arity galleries in which different candidates may have different
observed modality subsets.

% -----------------------------------------------------------------------------
\section{Additional Results and Analysis}
\label{sec:supp-additional}
% -----------------------------------------------------------------------------

\subsection{Components of Aggregation Gain}
\label{sec:supp-gain-components}

Table~\ref{tab:supp-gain-components} gives the two quantities used to compute
each aggregation-gain entry in Table~\ref{tab:gain}: the strongest
single-modality R@1 within a checkpoint and that checkpoint's multimodal
aggregation R@1. Values are rounded to one decimal place after evaluation.

ActivityNet and AudioCaps illustrate why the within-checkpoint decomposition
is useful. On ActivityNet, GRAM has a stronger best single pathway than
\ours{} ($36.8$ versus $34.1$), but its aggregated score is lower ($31.0$
versus $34.4$). AudioCaps shows a similar ordering: the best GRAM pathway is
$29.0$, compared with $25.7$ for \ours{}, while the aggregated scores are
$22.9$ and $27.1$, respectively. These cases suggest that the difference in
aggregate performance is not explained solely by stronger unimodal encoders;
the aggregation step materially affects the observed ordering.

\begin{table*}[t]
\centering
\caption{Components of aggregation gain: the best single-modality pathway
(``best''), the method's multimodal aggregate (``agg.''), and their difference
in text-to-video R@1. Superscripts follow the main paper.}
\label{tab:supp-gain-components}
\scriptsize
\setlength{\tabcolsep}{2.5pt}
\resizebox{\textwidth}{!}{%
\begin{tabular}{@{}l*{15}{c}@{}}
\toprule
& \multicolumn{3}{c}{MSR-VTT}
& \multicolumn{3}{c}{DiDeMo}
& \multicolumn{3}{c}{ActivityNet}
& \multicolumn{3}{c}{VATEX}
& \multicolumn{3}{c}{AudioCaps} \\
\cmidrule(lr){2-4}\cmidrule(lr){5-7}\cmidrule(lr){8-10}
\cmidrule(lr){11-13}\cmidrule(lr){14-16}
Method & best & agg. & gain & best & agg. & gain & best & agg. & gain
& best & agg. & gain & best & agg. & gain \\
\midrule
GRAM$^{\star}$
& 42.1 & 38.7 & $-3.4$ & 32.0 & 28.2 & $-3.8$
& 36.8 & 31.0 & $-5.8$ & 77.5 & 75.6 & $-1.9$
& 29.0 & 22.9 & $-6.1$ \\
PMRL$^{\star}$
& 43.7 & 31.5 & $-12.2$ & 38.3 & 28.7 & $-9.6$
& 39.8 & 30.0 & $-9.8$ & 81.2 & 53.6 & $-27.6$
& 32.0 & 33.4 & $+1.4$ \\
HyperGRAM$^{\dagger}$
& 42.5 & 39.1 & $-3.4$ & 32.1 & 32.0 & $-0.1$
& 36.3 & 34.3 & $-2.0$ & 78.0 & 78.0 & $+0.0$
& -- & -- & -- \\
\ours{}, uniform weights
& 41.1 & 34.1 & $-7.0$ & 34.5 & 27.0 & $-7.5$
& 30.0 & 26.0 & $-4.0$ & 80.3 & 69.1 & $-11.2$
& 28.6 & 28.8 & $+0.2$ \\
\textbf{\ours{}, query-weighted}
& 41.2 & 45.2 & $+4.0$ & 35.4 & 34.3 & $-1.1$
& 34.1 & 34.4 & $+0.3$ & 81.2 & 81.7 & $+0.5$
& 25.7 & 27.1 & $+1.4$ \\
\bottomrule
\end{tabular}%
}
\end{table*}
\subsection{How the Gramian Baseline Sees a Missing Modality}
\label{sec:maskedvol}

The missing-modality comparison is only meaningful if the Gramian
baseline is provided with a principled method to score an incomplete clip.
The construction is therefore stated explicitly.

The obvious implementation is to zero-fill: substitute $z_m = 0$ for the
absent modality and computing the volume without modification. This approach is problematic,
as it does not yield informative results. A zero vector introduces a zero row and
a zero column into the Gram matrix $G$, so $\det G = 0$ \emph{exactly},
for every masked clip regardless of its content. Every masked clip
receives the same constant score, the ranking among them is decided by
tie-breaking, and any resulting number would measure the tie-break rule
rather than the aggregator. We do not do this.

Instead, an absent modality is turned into an orthonormal phantom axis.
Writing $p \in \{0,1\}^{|M|+1}$ for the presence indicator (the query is
always present), the masked Gramian is
\begin{equation}
\tilde{G} = G \odot (p\,p^{\!\top}) + \operatorname{diag}(\mathbf{1} - p),
\label{eq:maskedgram}
\end{equation}
which zeroes the absent row and column and then writes a $1$ on its
diagonal. Expanding the determinant along that row gives
$\det \tilde{G} = 1 \cdot \det G_{P}$, where $G_{P}$ is the sub-Gramian
over the present modalities: the absent axis contributes a factor of
exactly one and the clip is scored at its own lower arity, with no
imputed content and no discarded candidate. When $p = \mathbf{1}$, the
construction reproduces the unmasked volume exactly, so the $r{=}0$
row of the table corresponds to the released baseline's protocol, unmodified.

The same procedure and presence rule ($\|z_m\| \le 0.5$, since
present embeddings are unit-norm and the loader zero-fills any modality it
cannot load) are applied during training, validation, and all evaluations, for
every method. Both identities are verified in the unit tests: that the
masked volume equals the directly computed lower-arity volume, and that
it is distinct from the degenerate zero-fill case.

Two important consequences should be noted. First, this treatment is
\emph{more} favorable than the released implementation supports, which
cannot represent an incomplete clip and instead discards it; the baseline is
therefore not disadvantaged by the evaluation protocol. Second, it leaves a genuine
asymmetry that the scores themselves cannot resolve: volumes at different
arities are not on a common scale, so a mixed-arity gallery is ranked according to
a quantity whose units vary across candidates, whereas \ours{} returns a
cosine in $[-1,1]$ at every arity. This is considered part of the
explanation for the gap observed in Table~3 rather than an artifact of it.
\subsection{Two-Stage Results Under Missing-Modality Masking}
\label{sec:supp-two-stage}

Table~\ref{tab:supp-missing-two-stage} reports the reranked R@1 values for the
same masked cells used in Table~\ref{tab:missing}, together with each
method's video-only cosine score at $r=90\%$. The first-stage table in the main
paper isolates the component directly modified by the aggregation rule. The
two-stage values here additionally reflect how each checkpoint-specific,
frozen reranker responds to inputs outside the complete-modality distribution
on which it was trained.

The separation between methods is generally smaller after reranking than at
the first stage, although the detailed ordering varies by dataset and masking
rate. On MSR-VTT at $r=90\%$, for example, the two-stage scores approach the
corresponding video-only cosine values: $33.2$ versus $32.1$ for \ours{} and
$32.9$ versus $32.8$ for GRAM. Larger differences remain on some other
benchmarks, showing that the reranker does not reduce to the video pathway in
every cell. The table therefore supports the stage-specific interpretation in
the main paper: robustness should be assessed at both the representation and
reranking stages.

\begin{table*}[t]
\centering
\caption{Two-stage text-to-video R@1 under the masks used for Table~3 of the
main paper. The final column gives each method's video-only cosine R@1 at
$r=90\%$. Bold marks the higher value within each dataset and rate; ties are
both bold.}
\label{tab:supp-missing-two-stage}
\small
\setlength{\tabcolsep}{6pt}
\begin{tabular}{@{}llrrrrrr@{}}
\toprule
Dataset & Method & 0\% & 25\% & 50\% & 75\% & 90\% & $\operatorname{cos}_{TV}$@90\% \\
\midrule
MSR-VTT & \textbf{\ours{}} & \textbf{54.8} & \textbf{47.8} & \textbf{41.1} & \textbf{37.0} & \textbf{33.2} & 32.1 \\
& GRAM$^{\star}$ & 52.5 & 47.0 & 40.9 & 35.4 & 32.9 & 32.8 \\
\rowcolor{gray!10}
& \emph{margin} & $+2.3$ & $+0.8$ & $+0.2$ & $+1.6$ & $+0.3$ & \\
\midrule
DiDeMo & \textbf{\ours{}} & \textbf{51.3} & 44.2 & 38.9 & 32.1 & 29.5 & 23.2 \\
& GRAM$^{\star}$ & 50.7 & \textbf{45.0} & \textbf{39.8} & \textbf{32.7} & \textbf{29.8} & 21.1 \\
\rowcolor{gray!10}
& \emph{margin} & $+0.6$ & $-0.8$ & $-0.9$ & $-0.6$ & $-0.3$ & \\
\midrule
ActivityNet & \textbf{\ours{}} & 55.8 & 47.9 & 40.0 & 33.8 & 30.7 & 23.7 \\
& GRAM$^{\star}$ & \textbf{56.3} & \textbf{48.3} & \textbf{41.4} & \textbf{35.5} & \textbf{32.3} & 24.8 \\
\rowcolor{gray!10}
& \emph{margin} & $-0.5$ & $-0.4$ & $-1.4$ & $-1.7$ & $-1.6$ & \\
\midrule
VATEX & \textbf{\ours{}} & \textbf{90.5} & \textbf{85.2} & \textbf{73.3} & \textbf{61.9} & 57.3 & 58.5 \\
& GRAM$^{\star}$ & 90.0 & 82.8 & 72.9 & \textbf{61.9} & \textbf{57.8} & 58.0 \\
\rowcolor{gray!10}
& \emph{margin} & $+0.5$ & $+2.4$ & $+0.4$ & $+0.0$ & $-0.5$ & \\
\midrule
AudioCaps & \textbf{\ours{}} & \textbf{35.2} & 24.6 & 19.2 & 13.2 & 9.9 & 10.2 \\
& GRAM$^{\star}$ & 32.2 & \textbf{26.0} & \textbf{21.0} & \textbf{15.5} & \textbf{11.4} & 9.5 \\
\rowcolor{gray!10}
& \emph{margin} & $+3.0$ & $-1.4$ & $-1.8$ & $-2.3$ & $-1.5$ & \\
\bottomrule
\end{tabular}
\end{table*}

\subsection{Modality-Subset Ladder}
\label{sec:supp-subset-ladder}

Table~\ref{tab:supp-subset-ladder} evaluates each checkpoint with one fixed
modality set for the entire gallery: $\{V\}$, $\{V,A\}$, or $\{V,A,S\}$.
This differs from the missing-modality experiment, in which clips are masked
independently and candidates of different arities compete in the same ranked
list. The fixed-subset ladder measures how the score changes as evidence is
added, but does not itself test cross-arity comparability.

On MSR-VTT, the \ours{} score increases from $41.2$ to $42.1$ and $45.2$ as
audio and subtitles are added, whereas the GRAM score decreases from $42.1$ to
$39.7$ and $38.7$. On VATEX, adding audio increases the \ours{} score from
$81.2$ to $81.9$, and adding subtitles changes it slightly to $81.7$; the full
aggregate remains $0.5$ above its video-only pathway. GRAM decreases from
$77.5$ to $75.6$ when audio is added and is unchanged when subtitles are added.
The $\{V\}$ row also verifies that a one-element aggregate reduces to the
text--video cosine for both scoring rules.

\begin{table}[t]
\centering
\caption{Modality-subset ladder using each method's first-stage aggregation
score (text-to-video R@1). The full $\{V,A,S\}$ value is the aggregate used in
the main paper's aggregation-gain analysis.}
\label{tab:supp-subset-ladder}
\small
\setlength{\tabcolsep}{4pt}
\begin{tabular}{@{}llrrrr@{}}
\toprule
Dataset & Method & $\{V\}$ & $\{V,A\}$ & $\{V,A,S\}$ & $\Delta_{V\rightarrow VAS}$ \\
\midrule
MSR-VTT & \textbf{\ours{}} & 41.2 & 42.1 & 45.2 & $+4.0$ \\
& GRAM$^{\star}$ & 42.1 & 39.7 & 38.7 & $-3.4$ \\
\midrule
VATEX & \textbf{\ours{}} & 81.2 & 81.9 & 81.7 & $+0.5$ \\
& GRAM$^{\star}$ & 77.5 & 75.6 & 75.6 & $-1.9$ \\
\bottomrule
\end{tabular}
\end{table}

\subsection{Latent-Space Analysis}
\label{sec:supp-latent}

\begin{figure*}[t]
\centering
\includegraphics[width=\textwidth,height=0.52\textheight,keepaspectratio]{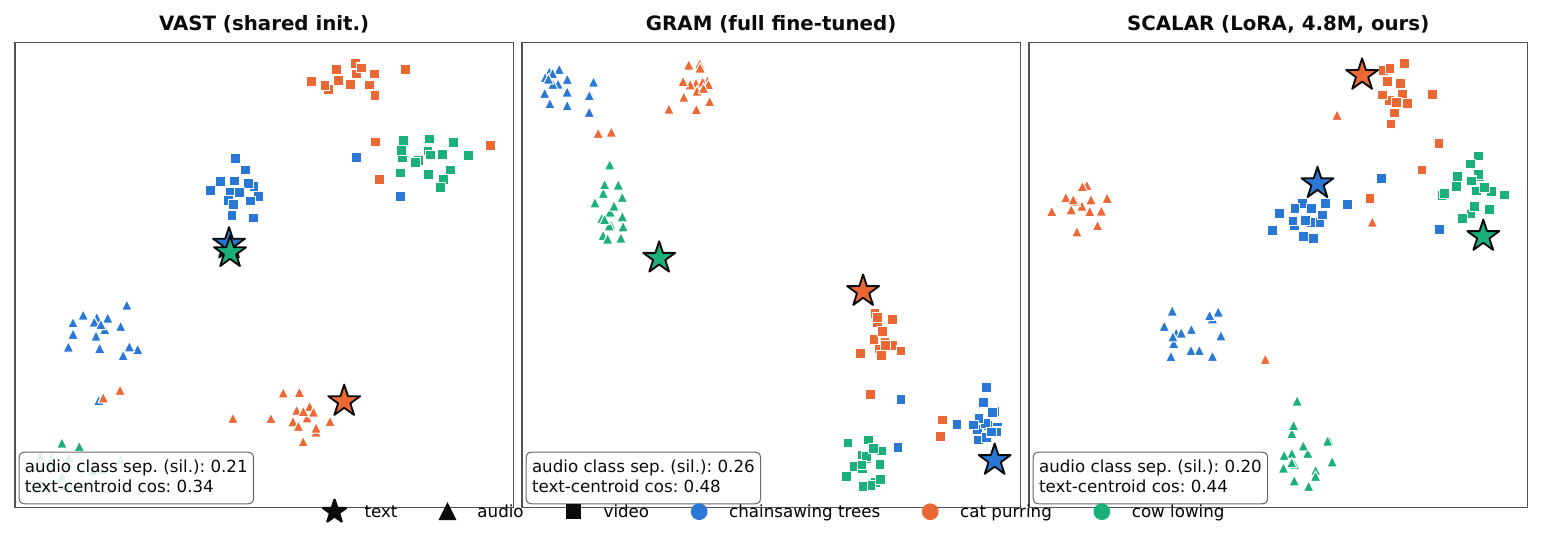}
\caption{Joint t-SNE of text, video, and audio embeddings on a VGGSound subset
for the shared VAST initialization, the fully fine-tuned GRAM checkpoint, and
\ours{}. Stars denote class-text embeddings, squares denote video, and
triangles denote audio. Panel annotations are computed in the original cosine
space over all eight classes, whereas three classes are displayed.}
\label{fig:supp-tsne}
\end{figure*}

Figure~\ref{fig:supp-tsne} compares embeddings from the shared VAST
initialization, the fully fine-tuned GRAM checkpoint, and \ours{}. The t-SNE
projection is used only for visualization. The numerical annotations are
computed in the original embedding space with cosine distance over all eight
dumped classes, rather than from the two-dimensional coordinates. For display,
we show the three classes with the largest improvement in text--centroid cosine
for both adapted models relative to the shared initialization.

Mean text--centroid cosine, computed between a class text embedding and the
mean of that class's video and audio embeddings, is $0.34$ for VAST, $0.48$ for
GRAM, and $0.44$ for \ours{}. Mean classwise audio silhouette is $0.21$,
$0.26$, and $0.20$, respectively. These measurements suggest that both adapted
models move the text anchors relative to the shared initialization. On these
particular diagnostics, \ours{} lies between or slightly below the shared
initialization and the fully fine-tuned GRAM checkpoint; the visualization is
therefore qualitative and is not used as evidence for the retrieval ranking.

\subsection{VGGSound Transfer and Protocol Check}
\label{sec:supp-vggsound}

Table~\ref{tab:supp-vggsound} reports audio-anchored
classification-as-retrieval on VGGSound-5K. \ours{} obtains
$34.9\pm0.3$ Acc@1, compared with $40.5$ for the released GRAM checkpoint.
The pathway decomposition localizes part of this difference: the \ours{}
video pathway is lower than GRAM's ($31.3$ versus $35.0$), whereas its audio
pathway is slightly higher ($27.0$ versus $26.3$). Its multimodal score is
$3.6$ points above its own stronger pathway, indicating positive aggregation
gain within the \ours{} checkpoint even though its final transfer result is
lower.

The uniform-weight control obtains $35.9$ Acc@1, exceeding the
query-conditioned configuration in this setting. VGGSound uses short class
labels rather than sentence queries, but the experiment does not isolate query
length from the change in task and domain. We therefore treat this result as a
boundary of the current evidence rather than attributing it to one factor.
Finally, the released GRAM checkpoint obtains $40.5$ Acc@1, within $0.1$ of its
published $40.6$, providing an additional protocol-consistency check on this
benchmark.

\begin{table}[t]
\centering
\caption{VGGSound-5K audio-anchored classification-as-retrieval using 310
class labels as queries over 5{,}000 clips. Results are Acc@1/Acc@10 without
reranking and are not directly comparable to the retrieval tables. The
\ours{} Acc@1 entry reports three seeds (mean $\pm$ standard deviation).}
\label{tab:supp-vggsound}
\small
\setlength{\tabcolsep}{5pt}
\begin{tabular}{@{}lcc@{}}
\toprule
Method & Acc@1 & Acc@10 \\
\midrule
GRAM (paper)$^{\S}$ & 40.6 & 78.1 \\
GRAM$^{\star}$ & 40.5 & 77.3 \\
\textbf{\ours{}} & $34.9\pm0.3$ & 71.2 \\
\midrule
\multicolumn{3}{@{}l}{\emph{Pathway decomposition} (\ours{} / GRAM$^{\star}$)} \\
\quad video pathway & 31.3 / 35.0 & 63.9 / 68.4 \\
\quad audio pathway & 27.0 / 26.3 & 63.3 / 66.4 \\
\quad \ours{}, uniform weights & 35.9 & 72.7 \\
\bottomrule
\end{tabular}
\end{table}

\subsection{Evaluation-Environment Audit}
\label{sec:supp-environment}

Table~\ref{tab:supp-environment} evaluates the same released GRAM checkpoint
in three environments. The measured MSR-VTT R@1 ranges from $52.5$ to $54.8$,
which is larger than the $0.2$ repeated-evaluation spread observed within our
environment. This comparison motivates the main paper's use of artifacts
evaluated under one environment, with published values shown only for
reference.

The offsets are not constant across datasets or methods. Relative to the
published results, our HyperGRAM training is lower by $2.6$, $2.4$, and $4.5$
R@1 on MSR-VTT, DiDeMo, and ActivityNet, respectively, but higher by $9.9$ on
the 431-clip VATEX gallery. The released GRAM checkpoint is lower by $2.3$ on
MSR-VTT and higher by $6.5$ on that VATEX gallery. These values should not be
read as a controlled estimate of one software component; they show why a
single constant correction would be inappropriate.

\begin{table}[t]
\centering
\caption{Evaluation-environment audit on MSR-VTT text-to-video-audio-subtitle
two-stage R@1. The first three rows use the same released GRAM checkpoint; the
last row is our GRAM retraining evaluated in our environment.}
\label{tab:supp-environment}
\small
\setlength{\tabcolsep}{4pt}
\begin{tabular}{@{}llr@{}}
\toprule
Weights & Evaluation environment & R@1 \\
\midrule
GRAM, released      & published & 54.8 \\
GRAM, released      & third-party reimplementation & 53.4 \\
GRAM, released      & ours      & 52.5 \\
GRAM, retrained by us & ours    & 52.4\,\tiny{$\pm$0.21} \\
\bottomrule
\end{tabular}
\end{table}

% \subsection{Statistical Support}
% \label{sec:supp-statistics}
\subsection{Statistical Support}
\label{sec:supp-statistics}

\paragraph{Missing-modality sweep.}
\ours{} exceeds GRAM in all 25 masking-rate--benchmark cells in the main
paper. Under an exact one-sided sign test with equal probability of either
ordering, this gives $p=2.98\times10^{-8}$. Because the five rates within a
benchmark share nested masks and are not independent, a conservative summary
that treats each benchmark as one unit is 5 of 5, giving $p=0.031$.

\paragraph{Aggregation gain.}
Twelve of the 14 measurable baseline method--benchmark cells have negative
aggregation gain. Under the same equal-probability sign-test null, the exact
one-sided value is $p=0.0065$. The method rows share checkpoints across
benchmarks, so the cell-level test should be interpreted descriptively rather
than as 14 fully independent trials.

\paragraph{Seed variation.}
Table~\ref{tab:supp-seed-ci} reports the mean, sample standard deviation, and a
two-sided 95\% $t$ interval for the three \ours{} seeds. GRAM and PMRL are
released checkpoints, and HyperGRAM is a single training run, so no comparable
baseline variance estimate or two-sample test is available. The intervals
therefore summarize variation across the three \ours{} runs only.

\begin{table}[t]
\centering
\caption{\ours{} seed statistics and single-run baseline results for two-stage
text-to-video R@1. Baselines are released checkpoints or one training run, so
only \ours{} has a variance estimate.}
\label{tab:supp-seed-ci}
\scriptsize
\setlength{\tabcolsep}{3pt}
\resizebox{\columnwidth}{!}{%
\begin{tabular}{@{}lccrrr@{}}
\toprule
Dataset & \ours{} mean $\pm$ sd & 95\% CI & GRAM$^{\star}$ & PMRL$^{\star}$ & HyperGRAM$^{\dagger}$ \\
\midrule
MSR-VTT & $54.6\pm0.2$ & $[54.1,55.2]$ & 52.5 & 54.3 & 54.0 \\
DiDeMo & $51.2\pm0.3$ & $[50.6,51.9]$ & 50.7 & 52.5 & 48.9 \\
ActivityNet & $55.9\pm0.2$ & $[55.6,56.3]$ & 56.3 & 54.1 & 53.7 \\
VATEX & $90.6\pm0.1$ & $[90.3,90.9]$ & 90.0 & 89.6 & 89.8 \\
AudioCaps & $35.0\pm0.4$ & $[34.0,36.0]$ & 32.2 & 34.4 & -- \\
\bottomrule
\end{tabular}%
}
\end{table}

\subsection{Trainable-Parameter Accounting}
\label{sec:supp-parameters}

We report an exact trainable-parameter count only for configurations that we
trained and instrumented. For a training run, this count is the number of
parameters passed to the optimizer. A released checkpoint, by contrast,
exposes floating-point tensors but does not record which tensors received
gradients. Treating every stored tensor as trainable would also include buffers
and task heads and is therefore not directly comparable.

Our full-fine-tuning control illustrates the difference. Counting parameters
updated by the optimizer gives $1{,}242{,}890{,}226$, whereas counting all
floating-point checkpoint tensors gives $1{,}397{,}367{,}145$, an 11\%
difference for the same model. A key-level comparison also finds a
$720{,}896$-parameter depth-projection head in our HyperGRAM and full-fine-
tuning checkpoints. The shared trunk constructs this head, but it is not used
by the modality configurations evaluated here; the released GRAM checkpoint
does not contain it. We therefore state the comparison supported by a common
definition: \ours{} updates $4{,}763{,}650$ parameters, whereas the baseline
methods use full-backbone pretraining, without assigning an artifact-derived
``trainable'' count to released checkpoints.

% -----------------------------------------------------------------------------
\section{Limitations}
\label{sec:supp-limitations}
% -----------------------------------------------------------------------------

\paragraph{Retrieval direction.}
The aggregation rule is explicitly conditioned on a text query and is designed
for text-to-video retrieval. In video-to-text retrieval, \ours{} remains
competitive but does not consistently match the strongest baseline, as
reported in the main paper. The benefits observed in one retrieval direction
should therefore not be assumed to transfer symmetrically.

\paragraph{Short label queries.}
On VGGSound, where the queries are class labels rather than sentences, the
uniform-weight control is stronger than the query-conditioned configuration
(Section~\ref{sec:supp-vggsound}). Because the task and domain also change,
this experiment does not by itself establish query length as the cause.

\paragraph{Backbone coverage.}
All experiments use VAST so that GRAM, PMRL, HyperGRAM, and \ours{} can be
compared within a shared backbone family. Whether the same aggregation-gain
patterns hold for other multimodal backbones remains to be evaluated.

\paragraph{Two-stage robustness.}
The first-stage separation between aggregators is often compressed after
reranking, particularly under missing-modality masking
(Section~\ref{sec:supp-two-stage}). The present work diagnoses this interaction
but does not train a reranker specifically for incomplete modality sets.

\paragraph{Continued-pretraining scale.}
Following the compared geometric-aggregation methods, continued pretraining
uses a nominal 150k-clip subset rather than the full VAST-27M corpus. The
conclusions therefore apply to the continued-pretraining regime evaluated here;
their behavior at substantially larger training scale is not established.

% If independent citations are added to the supplement, enable the following:
% {
%   \small
%   \bibliographystyle{ieeenat_fullname}
%   \bibliography{main}
% }

\end{document}